\documentclass{article}
\usepackage{ijcai24}

\usepackage{times}
\usepackage{soul}
\usepackage{url}
\usepackage[hidelinks]{hyperref}
\usepackage[utf8]{inputenc}
\usepackage[small]{caption}
\usepackage{graphicx}
\usepackage{amsmath}
\usepackage{amsthm}
\usepackage{booktabs}
\usepackage[switch]{lineno}

\usepackage[ruled,linesnumbered]{algorithm2e} 
\usepackage{balance}
\usepackage{subfig} 
\usepackage{amsfonts}
\usepackage{mathrsfs}
\usepackage{multirow}
\usepackage{makecell}
\usepackage{threeparttable}
\usepackage[normalem]{ulem}

\newcommand\eqcite[1]{Equation~\eqref{#1}}

\title{All in One: Multi-Task Prompting for Graph Neural Networks \\(Extended Abstract)\footnote{ Xiangguo Sun, Hong Cheng, Jia Li, Bo Liu, and Jihong Guan. All in one: Multi-task prompting for graph neural networks. In Proceedings of the 29th ACM SIGKDD Conference on Knowledge Discovery and 
Data Mining, pages 2120–2131, 2023. }}

\author{
Xiangguo Sun$^1$\and
Hong Cheng$^1$\and
Jia Li$^{2}$\and
Bo Liu$^3$\and
Jihong Guan$^4$
\affiliations
$^1$The Chinese University of Hong Kong\\
$^2$The Hong Kong University of Science and Technology (Guangzhou)\\
$^3$Southeast University, 
$^4$Tongji University
}

\begin{document}

\maketitle

\begin{abstract}
This paper is an extended abstract of our original work \cite{sun2023all} published in KDD23, where we won the best research paper award. 
The paper introduces a novel approach to bridging the gap between pre-trained graph models and the diverse tasks they're applied to, inspired by the success of prompt learning in NLP. Recognizing the challenge of aligning pre-trained models with varied graph tasks (node level, edge level, and graph level), which can lead to negative transfer and poor performance, we propose a multi-task prompting method for graphs. This method involves unifying graph and language prompt formats, enabling NLP's prompting strategies to be adapted for graph tasks. By analyzing the task space of graph applications, we reformulate problems to fit graph-level tasks and apply meta-learning to improve prompt initialization for multiple tasks. Experiments show our method's effectiveness in enhancing model performance across different graph tasks. 

\textit{Beyond the original work, in this extended abstract, we further discuss the graph prompt from a bigger picture and provide some of the latest work toward this area.}

\end{abstract}

\section{Introduction}

Graph neural networks (GNNs) \cite{sun2021heterogeneous} are increasingly applied across various fields \cite{sun2022self,sun2022your,li2024attention,sun2022structure,chen2020multi,sun2023counter}. The focus has shifted towards optimizing graph model training for specific problems. Traditional graph learning methods depend heavily on labels, often scarce or unfit for real-world complexities, leading to overfitting, especially with out-of-distribution data \cite{shen2021towards}. A popular mitigation strategy involves pre-training on accessible data, then fine-tuning for specific tasks \cite{jin2020self}, despite challenges in aligning pre-trained models with diverse downstream tasks.

A novel approach, inspired by NLP, combines pre-training with prompt learning and fine-tuning, where prompts facilitate task-specific model adjustments without extensive retraining. This method shows promise for efficient model adaptation, especially in scenarios with limited data. However, applying the concept of language prompts to GNNs introduces challenges, such as defining prompt content and integration with graph structures, and ensuring prompts effectively bridge pre-training tasks with varied downstream applications. Current efforts in graph prompt learning are limited and typically focus on single-task scenarios \cite{sun2022gppt}. We extend NLP prompt methods to GNNs for multi-task applications, addressing challenges in prompt design, task reformulation, and prompt optimization. Our contributions include a unified prompt format for language and graph domains, a strategy to reformulate tasks for better alignment with pre-training, and the application of meta-learning to enhance prompt efficacy across multiple tasks. Our extensive evaluations demonstrate the superiority of our approach.

\section{Motivations}

Graph pre-training \cite{sun2021multi} employs strategies to imbue GNNs with broad knowledge, reducing the need for task-specific annotations. Techniques vary from node and edge comparisons to graph-level contrastive learning, which proves superior in learning graph knowledge by enhancing graph representation or adjusting model parameters for consistency across perturbations \cite{you2020graph,xia2022simgrace,sun2022self}. Intuitively, the above graph-level pre-training strategies have some intrinsic similarities with the language-masked prediction task: aligning two graph views generated by node/edge/feature mask or other perturbations is very similar to predicting some vacant ``blanks'' on graphs. 
To this end, we aim to merge graph pre-training's depth with prompt learning's adaptability, addressing the multifaceted challenges in deploying GNNs across various tasks more effectively.

\section{Multi-task Prompting on Graphs}\label{sec:prompt}
This section presents a condensed overview of our approach to multi-task prompting for graph models, aiming to enhance the transferability of pre-trained graph models across various tasks without altering the original model architecture.

\noindent \textbf{Objective:} Our primary goal is to develop a graph prompt that seamlessly integrates with original graphs, thereby aligning pre-trained graph models more closely with diverse downstream tasks and improving knowledge transfer across domains.

\noindent \textbf{Framework Overview:} We introduce a multi-task prompting framework that first standardizes different graph tasks into a uniform format, focusing on graph-level tasks. We then design a novel graph prompt that incorporates learnable tokens, structures, and adaptive insertion patterns. To optimize the prompt for various tasks, we employ a meta-learning strategy, enabling the framework to adjust prompts dynamically for improved performance across multiple tasks.

\noindent \textbf{Reformulating Tasks for Generalization:} Recognizing the challenge of diverse task requirements in graphs, we reformulate node-level and edge-level tasks into graph-level tasks. This approach, inspired by the hierarchical nature of graph operations, allows for a broader application of pre-training knowledge by treating operations like node or edge modifications as graph-level changes.

\noindent \textbf{Designing the Prompt Graph:} We draw parallels between NLP and graph prompting, aiming for a unified representation that includes prompt tokens, token structures, and insertion patterns. This ensures that our graph prompts are both meaningful and adaptable to the structure of the original graph.

Let a graph instance be $\mathcal{G}=(\mathcal{V},\mathcal{E})$ where $\mathcal{V}=\{v_1,v_2,\cdots,v_N\}$ is the node set containing $N$ nodes; each node has a feature vector denoted by $\mathbf{x}_i \in \mathbb{R}^{1\times d}$ for node $v_i$; $\mathcal{E}=\{(v_i,v_j)|v_i,v_j \in \mathcal{V}\}$ is the edge set where each edge connects a pair of nodes in $\mathcal{V}$. With the previous discussion, we here present our prompt graph as $\mathcal{G}_p=(\mathcal{P},\mathcal{S})$ where $\mathcal{P}=\{p_1,p_2,\cdots,p_{|\mathcal{P}|}\}$ denotes the set of prompt tokens and $|\mathcal{P}|$ is the number of tokens. Each token $p_i \in \mathcal{P}$ can be represented by a token vector $\mathbf{p}_i\in \mathbb{R}^{1\times d}$ with the same size of node features in the input graph; Note that in practice, we usually have $|\mathcal{P}|\ll N$ and $|\mathcal{P}|\ll d_h$ where $d_h$ is the size of the hidden layer in the pre-trained graph model. With these token vectors, the input graph can be reformulated by adding the $j$-th token to graph node $v_i$ (e.g., $\mathbf{\hat{x}}_i= \mathbf{x}_i +\mathbf{p}_j $). Then, we replace the input features with the prompted features and send them to the pre-trained model for further processing.

$\mathcal{S}=\{(p_i,p_j)|p_i,p_j \in \mathcal{P}\}$ is the token structure denoted by pair-wise relations among tokens. Unlike the NLP prompt, the token structure in the prompt graph is usually implicit. To solve this problem, we propose three methods to design the prompt token structures: (1) the first way is to learn tunable parameters:
$$\mathcal{A}=\overset{|\mathcal{P}|-1}{\underset{i=1 \atop j=i+1}{\cup}}\{a_{ij}\}$$

\noindent where $a_{ij}$ is a tunable parameter indicating how possible the token $p_i$ and the token $p_j$ should be connected; (2) the second way is to use the dot product of each prompt token pair and prune them according to the dot value. In this case, $(p_i,p_j)\in \mathcal{S}$ iff $\sigma(\mathbf{p}_i\cdot \mathbf{p}_j)<\delta$ where $\sigma(\cdot)$ is a sigmoid function and $\delta$ is a pre-defined threshold; (3) the third way is to treat the tokens as independent and then we have $\mathcal{S}=\emptyset$.

Let $\psi$
be the inserting function that indicates how to add the prompt graph $\mathcal{G}_p$ to the input graph $\mathcal{G}$, then the manipulated graph can be denoted as $\mathcal{G}_m=\psi(\mathcal{G}, \mathcal{G}_p)$. We can define the inserting pattern as the dot product between prompt tokens and input graph nodes, and then use a tailored connection like
$\mathbf{\hat{x}}_i=\mathbf{x}_i+\sum_{k=1}^{|\mathcal{P}|} w_{ik}\mathbf{p}_k$ where $w_{ik}$ is a weighted value to prune unnecessary connections:
\begin{equation}
w_{ik}=\left\{
            \begin{array}{cl}
                 \sigma(\mathbf{p}_k\cdot \mathbf{x}_i^T), & \text{if } \sigma(\mathbf{p}_k\cdot \mathbf{x}_i^T)>\delta \\
                 0, & \text{otherwise}
            \end{array}
        \right.
\end{equation}
As an alternative and special case, we can also use a more simplified way to get $\mathbf{\hat{x}}_i=\mathbf{x}_i+\sum_{k=1}^{|\mathcal{P}|} \mathbf{p}_k$.

\noindent \textbf{Meta-Learning for Prompt Optimization:} We leverage meta-learning to refine our prompting approach, structuring the learning process to accommodate multiple tasks simultaneously. This method updates prompt parameters based on task-specific performances, ensuring that the final prompts are well-suited to a wide array of graph tasks.

\begin{figure}[t]
\centering
\subfloat[NLP tasks]{
\label{fig:hierarchy_2}
\includegraphics[width=0.24\textwidth]{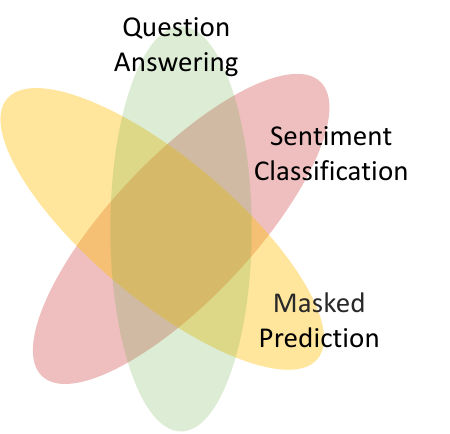}%
}
\subfloat[graph tasks]{
\label{fig:hierarchy_1}
\includegraphics[width=0.24\textwidth]{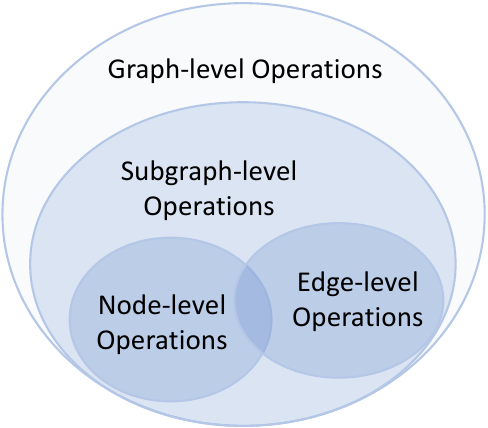}%
}
\caption{Task space in NLP and graph. Realizing the intrinsic nature of task space in the graph area, we reformulate node-level and edge-level tasks to graph-level tasks to achieve more general capabilities for graph models.}
\label{fig:hierarchy}
\end{figure}

\begin{figure}[h]
\centering
\subfloat[Induced graphs for nodes]{
\label{fig:ign}
\includegraphics[width=0.4\textwidth]{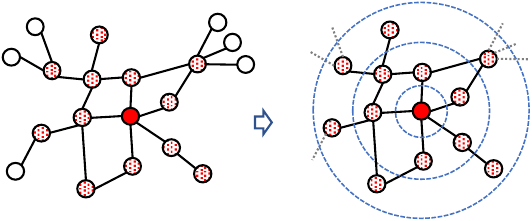}
}
\\
\subfloat[Induced graphs for edges]{
\label{fig:ige}
\includegraphics[width=0.4\textwidth]{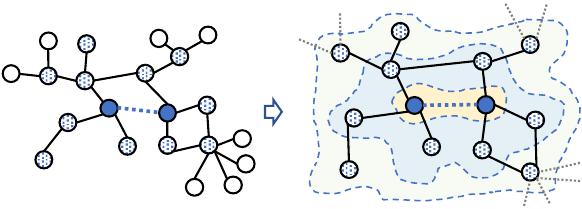}%
}
\caption{Reformulate node-level and edge-level tasks to graph-level tasks by induced graphs.}
\label{fig:ig}
\end{figure}

\begin{table*}[]
\centering
\caption{Node-level performance (\%) with 100-shot setting. IMP (\%): the average improvement of prompt over the rest.}
\label{tab:node_level}
\resizebox{0.8\textwidth}{!}{%
\begin{tabular}{@{}p{0.07\textwidth}<{\centering}c|p{0.02\textwidth}<{\centering}p{0.025\textwidth}<{\centering}p{0.035\textwidth}<{\centering}|p{0.025\textwidth}<{\centering}p{0.025\textwidth}<{\centering}p{0.035\textwidth}<{\centering}|p{0.025\textwidth}<{\centering}p{0.025\textwidth}<{\centering}p{0.035\textwidth}<{\centering}|p{0.025\textwidth}<{\centering}p{0.025\textwidth}<{\centering}p{0.035\textwidth}<{\centering}|p{0.025\textwidth}<{\centering}p{0.025\textwidth}<{\centering}p{0.035\textwidth}<{\centering}@{}}
\toprule
\multirow{2}{*}{\makecell[c]{Training\\ schemes}}     & \multirow{2}{*}{Methods} & \multicolumn{3}{c|}{Cora}       & \multicolumn{3}{c|}{CiteSeer}    & \multicolumn{3}{c|}{Reddit}     & \multicolumn{3}{c|}{Amazon} & \multicolumn{3}{c}{Pubmed}  \\
                                      &                          & Acc & F1 & AUC & Acc & F1 & AUC & Acc & F1 & AUC & Acc & F1 & AUC & Acc & F1 & AUC \\ \midrule
\multirow{3}{*}{supervised}           
& GAT                        & 74.45    & 73.21    & 82.97     & 83.00    & 83.20    & 89.33     & 55.64    & 62.03    & 65.38     & 79.00    & 73.42    & 97.81     & 75.00    & 77.56    & 79.72     \\
                                      & GCN                        & 77.55    & 77.45    & 83.71     & 88.00    & 81.79    & 94.79     & 54.38    & 52.47    & 56.82     & 95.36    & 93.99    & 96.23     & 53.64    & 66.67    & 69.89     \\
                                      & GT          & 74.25    & 75.21    & 82.04     & 86.33    & 85.62    & 90.13     & 61.50    & 61.38    & 65.56     & 85.50    & 86.01    & 93.01     & 51.50    & 67.34    & 71.91     
\\\midrule
\multirow{6}{*}{\makecell[c]{pre-train \\+\\  fine-tune}} 
& GraphCL+GAT                & 76.05    & 76.78    & 81.96     & 87.64    & 88.40    & 89.93     & 57.37    & 66.42    & 67.43     & 78.67    & 72.26    & 95.65     & 76.03    & 77.05    & 80.02     \\
                                      & GraphCL+GCN                & 78.75    & 79.13    & 84.90     & 87.49    & 89.36    & 90.25     & 55.00    & 65.52    & 74.65     & 96.00    & 95.92    & 98.33     & 69.37    & 70.00    & 74.74     \\
                                      & GraphCL+GT  & 73.80    & 74.12    & 82.77     & 88.50    & 88.92    & 91.25     & 63.50    & 66.06    & 68.04     & 94.39    & 93.62    & 96.97     & 75.00    & 78.45    & 75.05     \\
                                      & SimGRACE+GAT               & 76.85    & 77.48    & 83.37     & 90.50    & 91.00    & 91.56     & 56.59    & 65.47    & 67.77     & 84.50    & 84.73    & 89.69     & 72.50    & 68.21    & 81.97     \\
                                      & SimGRACE+GCN               & 77.20    & 76.39    & 83.13     & 83.50    & 84.21    & 93.22     & 58.00    & 55.81    & 56.93     & 95.00    & 94.50    & 98.03     & 77.50    & 75.71    & 87.53     \\
                                      & SimGRACE+GT & 77.40    & 78.11    & 82.95     & 87.50    & 87.05    & 91.85     & 66.00    & 69.95    & 70.03     & 79.00    & 73.42    & 97.58     & 70.50    & 73.30    & 74.22     
\\\midrule
\multirow{6}{*}{prompt}               
& GraphCL+GAT                & 76.50    & 77.26    & 82.99     & 88.00    & 90.52    & 91.82     & 57.84    & 67.02    & 75.33     & 80.01    & 75.62    & 97.96     & 77.50    & 78.26    & 83.02     \\
                                      & GraphCL+GCN                & 79.20    & 79.62    & 85.29     & 88.50    & 91.59    & 91.43     & 56.00    & 68.57    & 78.82     & 96.50    & 96.37    & 98.70     & 72.50    & 72.64    & 79.57     \\
                                      & GraphCL+GT  & 75.00    & 76.00    & 83.36     & 91.00    & 91.00    & 93.29     & 65.50    & 66.08    & 68.86     & 95.50    & 95.43    & 97.56     & 76.50    & 79.11    & 76.00     \\
                                      & SimGRACE+GAT               & 76.95    & 78.51    & 83.55     & 93.00    & 93.14    & 92.44     & 57.63    & 66.64    & 69.43     & 95.50    & 95.43    & 97.56     & 73.00    & 74.04    & 81.89     \\
                                      & SimGRACE+GCN               & 77.85    & 76.57    & 83.79     & 90.00    & 89.47    & 94.87     & 59.50    & 55.97    & 59.46     & 95.00    & 95.24    & 98.42     & 78.00    & 78.22    & 87.66     \\
                                      & SimGRACE+GT & 78.75    & 79.53    & 85.03     & 91.00    & 91.26    & 95.62     & 69.50    & 71.43    & 70.75     & 86.00    & 83.72    & 98.24     & 73.00    & 73.79    & 76.64  
\\ \midrule
\multicolumn{2}{c|}{IMP (\%)}                                        & 1.47     & 1.94     & 1.10      & 3.81     & 5.25     & 2.05      & 3.97     & 5.04     & 6.98      & 4.49&	5.84&	2.24  & 8.81     & 4.55     & 4.62      \\     
\midrule  
\multicolumn{2}{c|}{\makecell[c]{Reported Acc of GPPT (Label Ratio 50\%)}}        & 77.16    & --     & --      & 65.81     & --    & --     & 92.13     & --     & --      & 86.80  & -- & --  & 72.23    & --    & --      \\ \cline{1-2}
\multicolumn{2}{c|}{\makecell[c]{appr. Label Ratio of our 100-shot setting}}  & \multicolumn{3}{c|}{$\sim 25\% $}       & \multicolumn{3}{c|}{$\sim 18\% $}    & \multicolumn{3}{c|}{$\sim 1.7\% $}     & \multicolumn{3}{c|}{$\sim 7.3\% $} & \multicolumn{3}{c}{$\sim 1.5\% $}  \\
\bottomrule
\end{tabular}%
}
\end{table*}

\section{Why It Works?}\label{sec:why}

\noindent \textbf{Comparison to Prior Work:} While GPPT \cite{sun2022gppt} represents an early attempt at graph prompting, focusing on edge prediction for node classification, our method extends this concept significantly. Unlike GPPT, our framework is more versatile, accommodating a broader range of graph tasks and pre-training strategies beyond edge prediction, including advanced graph-level strategies like GraphCL \cite{you2020graph} and SimGRACE \cite{xia2022simgrace}.

\noindent \textbf{Flexibility:} Our approach introduces the concept of a prompt graph comprising multiple tokens with learnable structures, offering a more nuanced and flexible method for graph manipulation to better align with various pre-training strategies. We demonstrate that this flexibility allows for more effective adaptations of the graph structure to suit different tasks, reducing the error margin in representing manipulated graphs.

The nature of prompting is to manipulate the input data to match the pretext. Therefore, the flexibility of data operations is the bottleneck of prompting performance. Let $g$ be any graph-level transformation such as ``changing node features'', ``adding or removing edges/subgraphs'' etc., and $\varphi^{*}$ be the frozen pre-trained graph model. For any graph $\mathcal{G}$ with adjacency matrix $\mathbf{A}$ and node feature matrix $\mathbf{X}$, Fang et al. \cite{fang2022prompt} have proved that we can always learn an appropriate prompt token $p^*$ making the following equation stand:
\begin{equation}\label{equ:error_bound_naive}
    \varphi^*\left(\mathbf{A}, \mathbf{X}+p^*\right)=\varphi^*({g}(\mathbf{A}, \mathbf{X}))+O_{p\varphi}
\end{equation}
This means we can learn an appropriate token applied to the original graph to imitate any graph manipulation. Here {\small $O_{p\varphi}$} denotes the error bound between the manipulated graph and the prompting graph w.r.t. their representations from the pre-trained graph model. This error bound is related to some non-linear layers of the model (\textit{unchangeable}) and the quality of the learned prompt (\textit{changeable}), which is promising to be further narrowed down by a more advanced prompt scheme. In this paper, we extend the standalone token to a prompt graph that has multiple prompt tokens with learnable inner structures. Unlike the indiscriminate inserting in \eqcite{equ:error_bound_naive} (``{\footnotesize $\mathbf{X}+p^*$}'' means the prompt token should be added to every node of the original graph), the inserting pattern of our proposed prompt graph is highly customized. Let $\psi(\mathcal{G}, \mathcal{G}_p)$ denote the inserting pattern defined in section \ref{sec:prompt}; $\mathcal{G}$ is the original graph, and $\mathcal{G}_p$ is the prompt graph, then we can learn an optimal prompt graph $\mathcal{G}_p^*$ to extend \eqcite{equ:error_bound_naive} as follows:
\begin{equation}\label{equ:error_bound_new}
    \varphi^*\left(\psi(\mathcal{G}, \mathcal{G}_p^*)\right)=\varphi^*(\mathbf{g}(\mathbf{A}, \mathbf{X}))+O^{*}_{p\varphi}
\end{equation}
By efficient tuning, the new error bound {\small $O^{*}_{p\varphi}$} can be further reduced. That means our method supports more flexible transformations on graphs to match various pre-training strategies.


 \section{Evaluation}\label{sec:eva}

We compare our methods with other approaches on five public datasets including Cora \cite{welling2016semi}, CiteSeer \cite{welling2016semi}, Reddit \cite{hamilton2017inductive}, Amazon \cite{shchur2018pitfalls}, and Pubmed \cite{welling2016semi}. We compare our method with supervised, pre-training plus fine-tuning, and other prompt methods across node, edge, and graph-level tasks. Key findings include our method's superior performance in multi-task settings, showcasing notable improvements over existing methods.

\noindent \textbf{Multi-Task Performance}
Our study evaluates the performance of our prompt-based method across node-level, edge-level, and graph-level tasks in few-shot learning settings, comparing it against supervised methods and pre-training approaches. Results in Table \ref{tab:node_level} show that supervised methods struggle due to limited annotations available in few-shot scenarios, while pre-training methods offer better performance by leveraging prior knowledge. However, selecting and fine-tuning a pre-trained model for a specific task is effort-intensive and not always transferable to other tasks.

Our method, by incorporating prompts, shows compatibility improvements across all task levels, achieving performance boosts ranging from 1.10\% to 8.81\% for node-level tasks, 1.28\% to 12.26\% for edge-level tasks, and 0.14\% to 10.77\% for graph-level tasks. Notably, our approach under a more challenging setting (with only 100 labeled samples per class) still outperforms the GPPT model, which uses a 30\% to 50\% label ratio, indicating superior efficiency and adaptability of our method in few-shot learning contexts across various graph tasks. Please see the original paper for more task performance like edge-level and graph-level tasks.

\begin{table}[h]
\centering
\caption{Transferability (\%) on Amazon from different level tasks spaces. Source tasks: graph-level tasks and node-level tasks. Target task: edge-level tasks.}
\label{tab:trans_amazon}
\resizebox{0.4\textwidth}{!}{%
\begin{tabular}{@{}ll|lll@{}}
\toprule
Source task                  & Methods       & Accuracy & F1-score & AUC score \\ \midrule
\multirow{3}{*}{graph level} & hard & 51.50    & 65.96    & 40.34     \\
                             & fine-tune     & 62.50    & 70.59    & 53.91     \\
                             & prompt        & 70.50    & 71.22    & 74.02     \\ \midrule
\multirow{3}{*}{node level}  & hard & 40.50    & 11.85    & 29.48     \\
                             & fine-tune     & 46.00    & 54.24    & 37.26     \\
                             & prompt        & 59.50    & 68.73    & 55.90     \\ \bottomrule
\end{tabular}%
}
\end{table}

\noindent \textbf{Transferability:} Our method demonstrates enhanced adaptability, outperforming both hard transfer and fine-tuning approaches in transferring models to new tasks (as shown in Table \ref{tab:trans_amazon}) and domains (as shown in Table \ref{tab:trans_domains}). This is particularly evident in tasks requiring significant adaptation, where our prompting framework facilitates more effective knowledge transfer.

\noindent \textbf{Graph Transformation Flexibility:} Our approach effectively minimizes the error in representing manipulated graphs, demonstrating its capacity to support a wide range of graph transformations. This is further illustrated by visualizations that highlight the improved graph representations achieved through our prompting method.

For more experiments, please see in our original paper.

\begin{table}[h]
\centering
\caption{Transferability (\%) from different domains. Source domains: Amazon and PubMed. Target domain: Cora}
\label{tab:trans_domains}
\resizebox{0.4\textwidth}{!}{%
\begin{tabular}{@{}cc|ccc|ccc@{}}
\toprule
\multicolumn{2}{c|}{\makecell[c]{Source\\Domains}}    & \multicolumn{3}{c|}{Amazon}        & \multicolumn{3}{c}{PubMed}         \\ \midrule
Tasks                        &  & \makecell[c]{hard} & \makecell[c]{fine-tune} & prompt &\makecell[c]{hard} & \makecell[c]{fine-tune} & prompt  \\ \midrule
\multirow{3}{*}{\makecell[c]{node\\level}}  & Acc     & 26.9          & 64.14     & 65.07  & 55.62         & 57.93     & 62.07  \\
                             & F1      & 13.11         & 77.59     & 80.23  & 66.33         & 70.00     & 76.60  \\
                             & AUC     & 17.56         & 88.79     & 92.59  & 82.34         & 83.34     & 88.46  \\ \midrule
\multirow{3}{*}{\makecell[c]{edge\\level}}  & Acc     & 17.00         & 77.00     & 82.00  & 10.00         & 90.50     & 96.50  \\
                             & F1      & 10.51         & 81.58     & 84.62  & 2.17          & 89.73     & 91.80  \\
                             & AUC     & 4.26          & 94.27     & 96.19  & 6.15          & 93.89     & 94.70  \\ \midrule
\multirow{3}{*}{\makecell[c]{graph\\level}} & Acc     & 46.00         & 87.50     & 88.00  & 50.00         & 91.00     & 95.50  \\
                             & F1      & 62.76         & 89.11     & 88.12  & 10.00          & 93.90     & 95.60  \\
                             & AUC     & 54.23         & 86.33     & 94.99  & 90.85         & 91.47     & 98.47  \\ \bottomrule
\end{tabular}%
}
\end{table}

\begin{table}[h]
\centering 
\caption{Error bound discussed by section \ref{sec:why}
RED (\%): average reduction of each method to the original error.}
\label{tab:error}
\resizebox{0.45\textwidth}{!}{%
\begin{tabular}{@{}p{0.14\textwidth}<{\centering}c|ccc|c@{}}
\toprule
\makecell[c]{Prompt Solutions} & \makecell[c]{Token\\ Number} & \makecell[c]{Drop \\Nodes} & \makecell[c]{Drop \\Edges} & \makecell[c]{Mask \\Features} & RED (\%) \\ \midrule
\makecell[c]{Original Error \\(without prompt)} & 0        & 0.9917    & 2.6330    & 6.8209  &  -   \\\midrule
\makecell[c]{Naive Prompt \\(Equation \ref{equ:error_bound_naive})} & 1       & 0.8710    & 0.5241    & 2.0835  &  66.70$\downarrow$   \\\midrule
\multirow{3}{*}{\makecell[c]{Our Prompt Graph\\ (with token, structure,\\ and inserting patterns)}} & 3         & 0.0875    & 0.2337    & 0.6542  & 90.66$\downarrow$    \\
& 5         & 0.0685    & 0.1513    & 0.4372  & 93.71$\downarrow$    \\
& 10        & 0.0859    & 0.1144    & 0.2600  & 95.59$\downarrow$    \\ \bottomrule
\end{tabular}%
}
\end{table}



\section{A Bigger Picture of Graph Prompts}

In the rapidly evolving field of Artificial General Intelligence (AGI) \cite{li2024survey}, significant advancements have been made, especially with applications like ChatGPT in NLP and Midjourney in Computer Vision (CV), greatly enhancing our efficiency and creativity. Yet, the application of AGI in graph data analysis remains nascent, despite its potential to revolutionize areas such as drug design and battery development due to challenges in harmonizing information across modalities, domains, and tasks.

In section \ref{sec:why}, we can find that graph prompt has the potential to simulate various data manipulations. This means it can be used to achieve the tough challenge in graph domain transfer. Our paper also demonstrates its huge potential for task transfer. In addition, since we only need to tune a light-weight prompt while keeping a large graph model unchanged, it is more efficient. Prompt learning emerges as a promising solution. It has shown remarkable success in NLP and CV by reformulating tasks to leverage pre-trained models without extensive tuning. Prompt learning's efficiency in knowledge extraction and task reformulation presents an opportunity to address the complexities of working with graph data, suggesting a path to extend its benefits to graph-based AGI applications.

To achieve this end, some recent works have been proposed to follow up our paper \cite{sun2023graph}. We recently further studied the feasibility of domain transferring with graph prompt \cite{zhao2024all}, explored the application of graph prompt in protein multimer structure prediction \cite{gao2024protein}, and proposed various variants of graph prompt \cite{chen2024prompt}. We also release ``ProG'' (Prompt Graph), which is a Python library built upon PyTorch to easily conduct single or multi-task prompting for pre-trained Graph Neural Networks (GNNs). Please use the library at \url{https://github.com/sheldonresearch/ProG/}

In future work, we can further study the integration of graph prompts with various graph models \cite{zhang2022graph}, extend its applications\cite{liu2022nowhere,piao2023computing,cui2023dynamic,meng2023recognize}, and the security issue of graph prompt\cite{zhang2022unsupervised,yang2023generating}.

\section*{Acknowledgments}

This research is supported in part by project \#MMT-p2-23 of the Shun Hing Institute of Advanced Engineering, The Chinese University of Hong Kong, by grants from the Research Grant Council of the Hong Kong Special Administrative Region, China (No. CUHK 14217622), NSFC (No. 61972087, No. 62206067, No. U1936205, No. 62172300, No. 62202336), Guangzhou-HKUST(GZ) Joint Funding Scheme (No. 2023A03J0673), National Key R\&D Program of China (No. 2022YFB3104300, No. 2021YFC3300300), the Fundamental Research Funds for the Central Universities (No. ZD-21-202101), and Open Research Projects of Zhejiang Lab (No. 2021KH0AB04). \textit{The first author, Dr. Xiangguo Sun, in particular, wants to thank his parents for their kind support during his tough period.}

\clearpage

\bibliographystyle{named}
\bibliography{ijcai24}

\end{document}